\documentclass[preprint,12pt]{elsarticle}

\usepackage{amssymb}

\begin{document}

\begin{frontmatter}

\title{Enhancing Answer Selection in Community Question Answering with Pre-trained and Large Language Models}

\author[author1]{Xinghang Hu}
\affiliation[label1]{organization={PricewaterhouseCoopers},
            country={China}}



\begin{abstract}
Community Question Answering (CQA) becomes increasingly prevalent in recent years. However, there are a large number of answers, which is difficult for users to select the relevant answers. Therefore, answer selection is a very significant subtask of CQA. In this paper, we first propose the Question-Answer cross attention networks (QAN) with pre-trained models for answer selection and utilize large language model (LLM) to perform answer selection with knowledge augmentation. 
Specifically, we apply the BERT model as the encoder layer to do pre-training for question subjects, question bodies and answers, respectively, then the cross attention mechanism selects the most relevant answer for different questions. 
Experiments show that the QAN model achieves state-of-the-art performance on two datasets, SemEval2015 and SemEval2017.
Moreover, we use the LLM to generate external knowledge from questions and correct answers to achieve knowledge augmentation for the answer selection task by LLM, while optimizing the prompt of LLM in different aspects. The results show that the introduction of external knowledge can improve the correct answer selection rate of LLM on datasets SemEval2015 and SemEval2017. Meanwhile, LLM can also select the correct answer on more questions by optimized prompt.
\end{abstract}




\end{frontmatter}


\section{Introduction}

\begin{figure}
    \centering
    \includegraphics[width=0.8\linewidth]{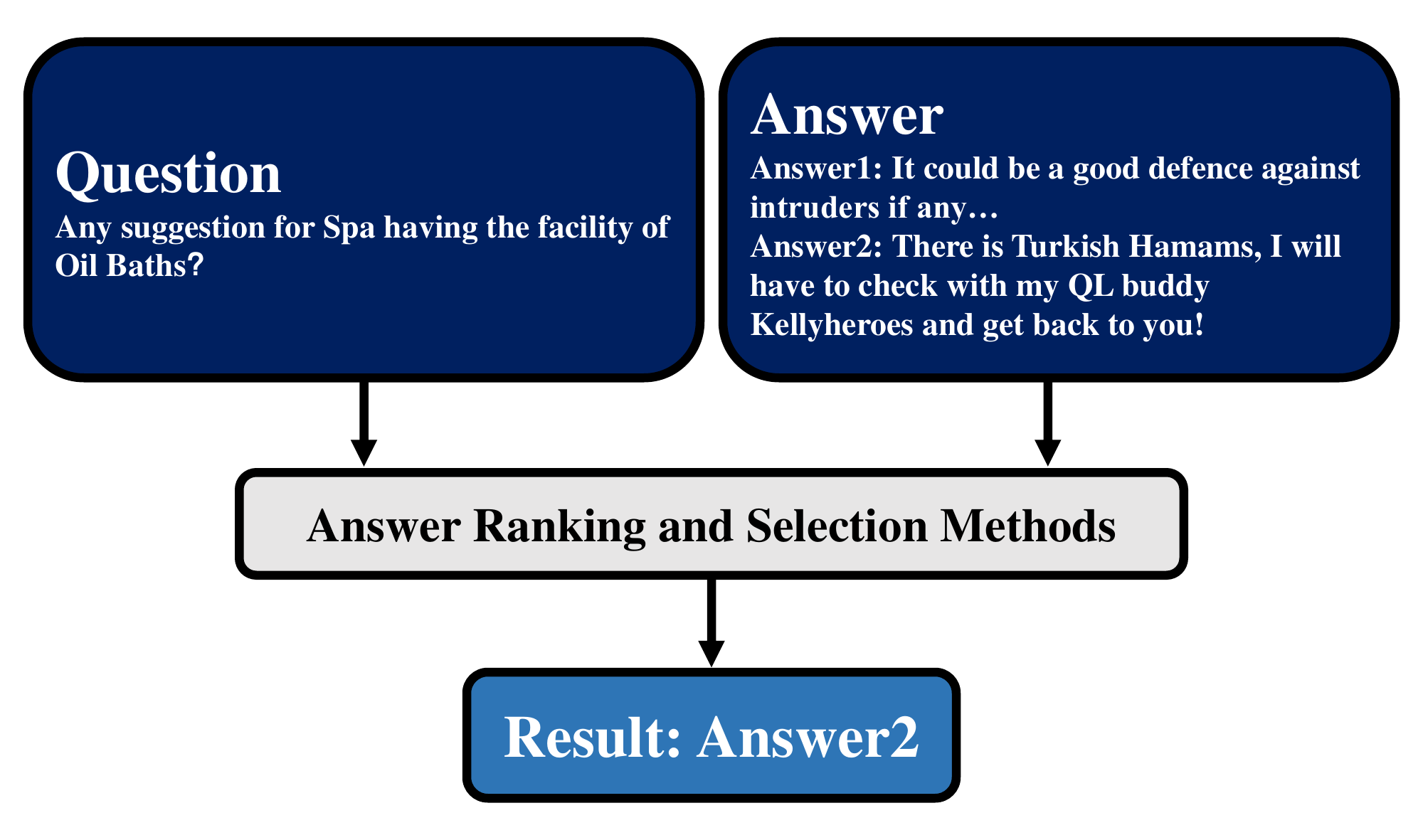}
    \caption{A schematic of a community question answering system}
    \label{fig:schemic}
\end{figure}

Community Question Answering (CQA), as shown in Figure~\ref{fig:schemic}, has gained popularity in recent years, offering an interactive experience and faster information retrieval across various domains. For instance, according to the latest available data, Quora boasts approximately 190 million active monthly users. Additionally, the total number of users visiting Quora, including unregistered users, reaches approximately 633 million per month, with 300 million being unique visitors. CQA platforms like Yahoo! Answers, StackOverflow, and Quora provide users with an open environment to search for information of interest, post their own inquiries, and contribute answers based on their knowledge.

However, due to the diverse backgrounds of the participants, the quality of answers can vary significantly. Some responses align with the topic and meet the users' expectations, while others deviate completely from the intended purpose. Consequently, users face the challenge of sifting through numerous candidate answers to find the most relevant one, which can be time-consuming. Furthermore, the number of repeated and unanswered questions has significantly increased. Many of these questions have, in fact, been previously asked and answered by different users but with slight variations in wording or syntax. Therefore, answer selection becomes a crucial task in CQA. 

The objective of answer selection is to identify the most pertinent answer from the repository, effectively reducing the time users spend reviewing all candidate responses. This not only enhances the user experience but also enables the resolution of similar, unresolved questions that have been answered before but presented differently.
By improving answer selection in CQA, users can enjoy a more efficient and satisfying experience while benefiting from the accumulated knowledge within the community. But the task of answer selection poses significant challenges for two main reasons. Firstly, it is complicated and tricky due to the presence of low-information words, auxiliary verbs, a large number of synonyms, and syntactic transformations. This complexity can be observed in examples such as the questions ``What are better ways to look for nice local restaurants?'' and ``How do you search for great restaurants along your route?'' which essentially convey the same meaning of seeking good places to eat, but differ in their choice of words and syntactic structures. Secondly, many researchers tend to treat questions and answers equally, overlooking the redundancies and noise present in answers, as pointed out by Zhang et al \cite{Zhang2017AttentiveIN}. This disregard for the informative content in answers often leads to significant deviations.
Answer selection provides an effective solution to address these challenges. It involves selecting valuable answers from a list of candidate responses based on the semantic matching between the question and the answer. Traditional approaches to answer selection typically rely on recurrent neural networks (RNN), Long Short-Term Memory (LSTM) recurrent neural networks, Seq2seq models, and other related technologies \cite{roy2023analysis,zhang2020novel}. With the emergence of attention mechanisms, such as the Transformer and BERT technologies, utilizing attention for answer selection has become a prominent research focus in the field of CQA \cite{Zhang2021GraphBasedTN,8736313,ha2020supervised}.


Nevertheless, current research in this area often neglects the separate processing of questions and answers, failing to adequately address the interaction between them. To address these challenges, we present QAN, a Question-Answer Cross Attention Network with Pre-trained models for Answer Selection in CQA. QAN leverages BERT, a pre-trained language model, for word encoding, taking into account the contextual information of each word in question subjects, question bodies, and answers. We employ cross attention between words in question subjects and words in answers, as well as between words in question bodies and words in answers, to capture interactive information between questions and answers. Finally, the QAN model combines attention on questions and attention on answers to compute the matching probability of question-answer pairs, enabling the identification of the most suitable answer for a given question.


Large Language Models(LLM) are deep learning models with massive parameters used for natural language processing tasks, capable of understanding context, generating text, and finding extensive applications across various domains. In the field of question answering, users can ask questions to an intelligent question answering system, and the LLM can help the system understand the question and provide relevant suggestions such as in medicine \cite{singhal2023large,singhal2023towards,lievin2022can} and materials science \cite{zaki2023mascqa}. In addition, LLM can also be used as an external knowledge engine for question answering systems \cite{shao2023prompting}. Although LLM has been studied for Q\&A tasks, few have been used for CQA, let alone for answer selection based on augmentation of external knowledge. Meanwhile, Prompt has an important effect on the output of large language models, but the related research is also less. To address these challenges, we use the external knowledge generated by the large language model LLaMa to enhance the performance of the answer selection task performed by LLM, and optimize the prompt input to LLM in different aspects to help select the correct answer from more questions.


The key contributions of our work are as follows:

\begin{itemize}
    \item We utilize the BERT model for pre-training question subjects, question bodies, and answers separately, capturing comprehensive semantic information from both questions and answers. Additionally, we employ the cross attention mechanism to effectively capture important interactive features between questions and answers. Our proposed model achieves state-of-the-art performance on the SemEval2015 and SemEval2017 datasets, outperforming existing answer selection models.
    \item We incorporate llama-7b-hf, a large language model, to generate knowledge as the answer reference for questions of the datasets. This reference enhances the alignment between questions and answers and improves the model's performance on the answer selection task.
    \item We optimized prompt from four different perspectives, enabling LLM to select the right answers on more questions of the datasets, providing ideas and direction for prompt optimization.
\end{itemize}

\section{Related Work}

Traditional methods for answer selection can be categorized into content/user modeling methods and adaptive support methods. Content/user modeling methods focus on modeling user characteristics, questions, and corresponding answers to extract high-level attributes from low-level Q\&A interactions, which are essential inputs for CQA functions. For instance, answer quality evaluation outputs can be utilized to rank answers. Shah \cite{shah2010evaluating} et al. constructed an improved evaluation system by extracting various features from questions, answers, and the users who posted them. Adaptive support methods, based on content/user modeling, enhance user collaboration success and effectiveness through question retrieval and question routing.
Question retrieval recommends archived question-answer pairs. Zhang  et al. \cite{zhang2014question} proposed a supervised question answering topic modeling method that matches questions not only at the term level but also at the topic level, demonstrating strong retrieval performance in CQA. Question routing recommends the best potential answer, taking into account user expertise, user activity, and motivation. Zhao \cite{zhao2014expert} approached the routing problem from the perspective of missing value estimation and used graph regularization matrix completion to estimate the missing values.

Attention mechanisms \cite{vaswani2017attention} have been widely adopted in question answering tasks. Guo et al. \cite{guo2021re} proposed a reattention framework for visual question answering tasks. Zheng et al. \cite{zheng2021mutual} incorporated attention mechanisms and bilinear technology to enhance features and address remote sensing visual question and answer tasks. Zhang et al.  \cite{Zhang2021GraphBasedTN} designed a graph-based three-attention network to construct target-aware responder representations, answer-specific question representations, and context-aware answer representations through attentional computation.
Ha et al. \cite{ha2020supervised} integrated supervised attention into match-LSTM, guiding attention weight learning for question-answer pairs with external lexical semantics, resulting in better performance compared to the baseline model. Cross attention within the attention mechanism is particularly adept at capturing interactive features between processing questions and answers due to its ability to handle cross-sequence relationships.

Large language models (LLM), exemplified by chatGPT, have gained significant attention in natural language processing. There have been some researches on large language models for Q\&A tasks. Guo et al. \cite{guo2022medical} focused on building a retrieval-based medical question answering system using large language models, semantic matching, named entity recognition, and knowledge graphs to improve answer selection, achieving consistent improvements over strong baselines on various datasets. Singhal et al. \cite{singhal2023large} presented MultiMedQA, a comprehensive benchmark for medical question answering, evaluates large language models, including Flan-PaLM with instruction prompt tuning, achieving state-of-the-art accuracy but still existing gaps compared to clinicians. Yu et al.  \cite{yu2022generate} used large language model generators that replace document retrievers to perform open-domain question answering, achieving significant performance improvements without external document retrieval. Lee et al. \cite{lee2023towards} introduced automated feedback, and implemented a feedback learning loop to improve citation, correctness, and fluency in LLM-generated responses for QA systems. Weng et al. \cite{weng2023large} introduced the Holistically Thought (HoT) method for enhancing large language models in addressing complex medical conversational question answering tasks, and a better response effect has been achieved. It can be seen that LLM have a broad application prospect in Q\&A systems, but there are still few studies on CQA tasks.

In recent years, the research of knowledge augmentation for question answering system has become a hot topic. Cai et al. \cite{10.1145/2063576.2063768} used the semantics of Wikipedia to enrich questions and solve the task of large-scale question classification. Jing et al. \cite{jing2022knowledge} proposed a knowledge-enhanced attentionanswer selection (KAAS) model to improve the performance of CQA systems. Huang et al. \cite{huang2020interactive} proposed an interactive knowledge-augmented attention network for answer selection (IKAAS), which uses the external knowledge in the knowledge graph as learning information for question answering pairs, and achieves good results. Li et al. \cite{li2023kepr} developed a well-performing answer ranking model by extending questions with wiki commonsense knowledge and capturing relevant knowledge. Nguyen et al. \cite{nguyen2021model} combined features from convolutional neural networks and other methods to obtain additional knowledge to enhance deep learning models and achieved better results in CQA tasks. It can be seen that knowledge enhancement is helpful to improve the correct rate of answer selection, so we also combine external knowledge and questions and answers to perform CQA tasks.

\section{Methodology for PLM}
In this section, we adopt PLM to make answer selection.
We introduce our proposed QAN model, including the mathematical expression and the components of the model.

\subsection{Task Description}

In this research, the answer selection task in CQA can be described as a tuple of four elements $(S, B, A, y)$. $S=[s^1,s^2,\cdots,s^m]$ represents a question subject whose length is $m$. $B=[b^1,b^2,\cdots,b^g]$ represents the corresponding question body whose length is $g$. $A=[a^1,a^2,\cdots,a^n]$ represents the corresponding answer whose length is n. And $y\in Y$ represents the relevance degree, namely, $y=\{Good, Potential, Bad\}$ to determine whether a candidate can answer a question properly or not. More detailed, Good represents that the answer can provide a proper solution for the question, while Potential indicates that the answer might provide a useful solution to users and Bad means the answer is not relevant to the question. Generally, our QAN model on the answer selection task in CQA can be summarized as assigning a label to each answer based on the conditional probability $Pr(y\mid S, B, A)$ with the given set $\{S, B, A\}$.

\subsection{Overview of Proposed Model}


\begin{figure}
    \centering
    \includegraphics[width=0.8\textwidth]{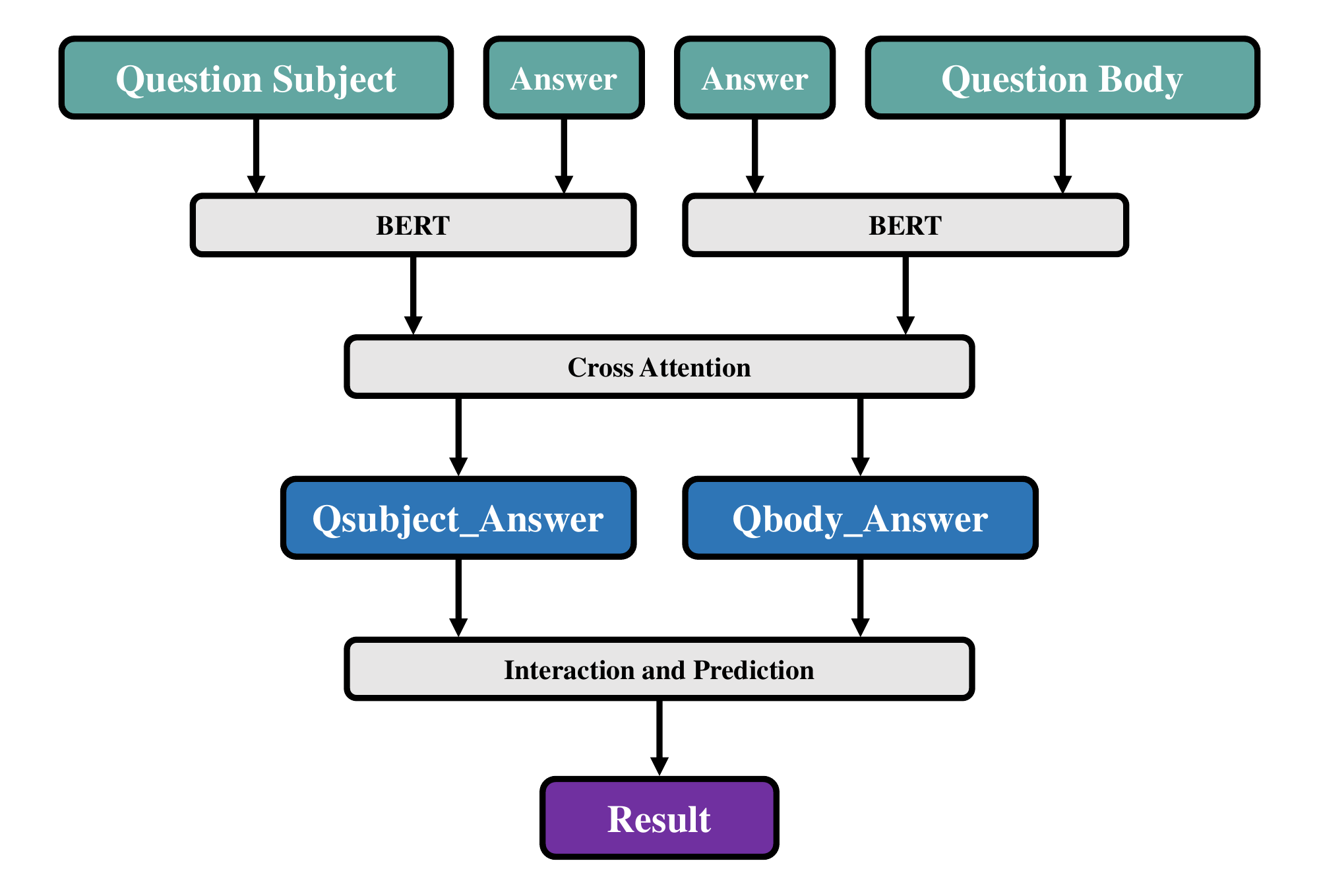}
    \caption{Pipeline of answer selection with PLMs.} 
    \label{fig:QAN}
\end{figure}

Our proposed QAN model consists of three key layers and the framework is shown in Figure~\ref{fig:QAN}.
We first utilize the BERT \cite{devlin2019bert} to capture contextual representations of the question subject, question body, and answer in token form. 
Next, we use cross attention mechanism \cite{Lin2017ASS} to establish essential interaction features between the question and answer by analyzing the relationships between the question subject-answer and question body-answer.
Specifically, this Cross Attention Layer takes encoded form ouputs 
by BERT encoder as input and compute the relevance between each word in question subjects and answers as well as the counterpart in question bodies and answers.
Take the process of computing relevance between each word in question subjects and answers as an example, we get a matrix. 
Then two similarity matrices, question subjects to answers and answers to question subjects,
are generated after normalization over each row and column by softmax function of the similarity matrix.
In similarity matrices, rows represent question subjects, and columns represent answers. And we get interactive features between question subjects and answers
as the final output.
After that, we propose the Interaction and Prediction Layer, inspired by previous work \cite{Chen2017,Mou2016}, processes interaction features and leverages acquired knowledge to assign labels to each answer in response to the given question, using conditional probability calculations. Specifically, we train our model with bidirectional GRU (Bi-GRU) to acquire context information between questions and their corresponding answers. After that, we use max pooling and mean pooling of quesitons and answers to acquire fixed-length vectors. Then, we concatenate these vectors to get the global representation $r$. Finally, we pass the global representation $r$ to the prediction layer which is consisted of a multi-layer perceptron (MLP) classifier to determine whether the semantic meaning of the give question-answer pair is equivalent or not.
%
%
%
These layers are hierarchically organized, with Figure 1 depicting the BERTCAN model's workflow from top to bottom.

\section{Methodology for LLM}
In this section, we adopt LLM for answer selection.
It includes knowledge retrieval, knowledge augmentation and prompt optimization.
We use Llama-7b-hf to generate external knowledge and perform answer selection. Llama-7b-hf is a collection of pretrained and fine-tuned generative text models ranging in scale of 7 billion parameters.

\subsection{Task Description}
\begin{figure}
\centering
\includegraphics[width=0.7\textwidth]{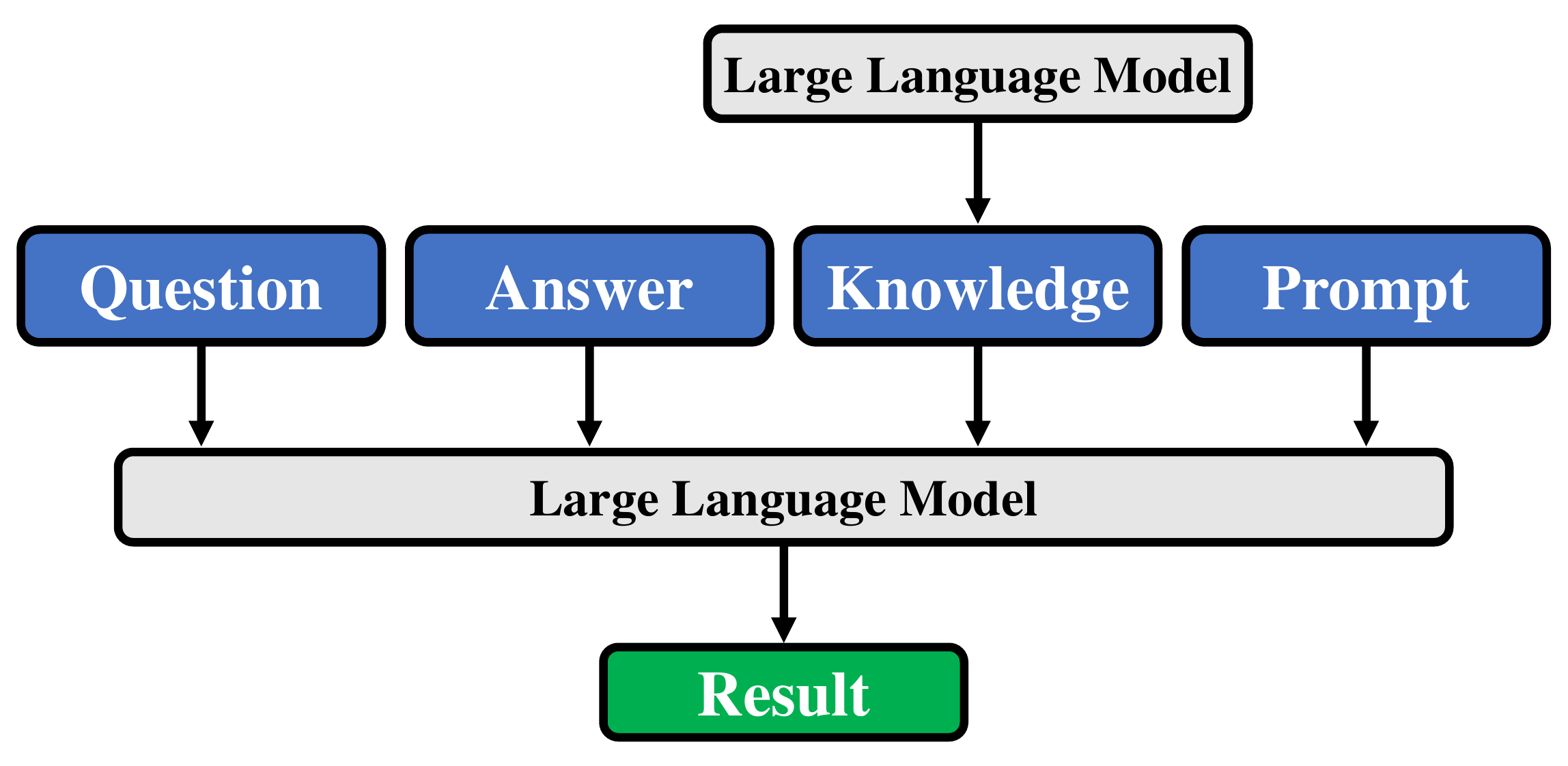}
\caption{Pipeline of answer selection with LLMs.} 
\label{llm}
\end{figure}

The BERTCAN model we proposed is used for answer multi-classification tasks. However, for LLM, their multi-classification task capabilities are not ideal \cite{balikas-2023-john}, so the task of LLM is changed to generating answer selection questions, that is, selecting the correct answers to the questions. In the SemEval2015 and SemEval2017 data sets, the correct answer is the answer where its ``CGold'' is ``Good''. The whole process is shown in Figure~\ref{llm}. In this research, the answer selection task in CQA by LLM can be described as a tuple of six elements $(S, B, A, K, P, y)$. $S$ represents a question subject. $B$ represents the corresponding question body. $A$ represents the corresponding answers. $K$ represents the corresponding knowledge to the question. $P$ represents the prompt that is input to LLM. And $y$ represents the output of the LLM.
Specifically, we combine question subject, question body, answer, knowledge and prompt, and then input the combined content into the LLM. LLM will output what it thinks is the most appropriate answer based on the input.

\subsection{Prompt Optimization}
Prompt has a crucial impact on the output of LLM. For the same question, different prompts may lead to different answers, as shown in Table~\ref{tab: differ prompt}. From the Table~\ref{tab: differ prompt}, we can see that not all prompts could select the right answer from the question that one prompt could not lead to the right option while another prompt could. Because Prompt2 is optimized compared to Prompt1. Consequently, the optimization of prompts on LLM is a key approach to improve the performance of answer select of CQA by LLM. In order to improve the accuracy of answer selection as much as possible, Prompt needs to be optimized. We optimize Prompt in terms of prompt length, position of questions and answers, question subject, position of task description.

 \begin{table*}[!ht]\tiny
    \centering
    \caption{An example where different prompts result in different results}
    \begin{tabular}{p{2.5cm}p{1.5cm}p{1cm}p{1.5cm}p{2.5cm}p{1cm}p{1cm}}
    \hline 
    \textbf{Question} & \textbf{Options} & \textbf{Right Opiton} & \textbf{Prompt1}&\textbf{Prompt2} & \textbf{Selection of Prompt1} &\textbf{Selection of Prompt2}\\ \hline
    Recently Indian Government bought 200 Tons of Gold from other country....The Gold price suddenly increase to a tremendously high. Why Indian so crazy in wearing Gold or golden accessories...their men and women wearing gold or golden jewelry; accessories extraordinarily...friend of mine wearing golden rings on 3 of his finger right and left...it's insane...
    &C1:It's a safe investment and they have been doing it for a long time.

    C2:And your concern about it is? One life to live; live it to the fullest.
    &C1
    &Utilizing the information in hint-[KNOWLEDGE], choose the serial number of the optimal response to the question-[QUESTION] that is distinct from the options-[ANSWER].
    &Given the discoveries from hint-[KNOWLEDGE], your task is to discern and specify the unique serial number tied to the most appropriate answer to question-[QUESTION]. Keep in mind that this response must be distinctive and one of the existing options listed in options-[ANSWER]. This involves analysis, cognition, and the capability to pinpoint distinctly pertinent information.
    &C2
    &C1
       \\ \hline
    \end{tabular} 
    \label{tab: differ prompt}
    \end{table*}

\subsection{Knowledge Retrieval}
Related studies have shown that knowledge augmentation, such as incorporating external knowledge graph, is helpful for improving the performance of question answering tasks ~\cite{jing2022knowledge, zhou2020multidomain,bian2021benchmarking}. Considering that external knowledge enhancement may be beneficial to improving the accuracy of answer choices, we combine the text of the question and the Good answer to the question, and use LLM to generate knowledge that helps select the correct answer. The following Table~\ref{tab:k gene} is an example of combining questions and answers to generate knowledge.

\begin{table*}[!ht]\tiny
    \centering
    \caption{An example of using LLM to generate knowledge about a relevant question.}
    \begin{tabular}{p{5cm}p{1.5cm}p{2.5cm}p{3cm}}
    \hline
        \textbf{Prompt} & \textbf{Quesiton}  & \textbf{Good Answer} & \textbf{Output} \\ \hline
        From the provided question [QUESTION] and answer [GOOD ANSWER], please generate a short piece of related knowledge. Your response should be concise and provide relevant information that pertains to the question. Feel free to draw from various sources and provide interesting and educational insights related to the question.&Any good place to shop? Gucci?Thanks for your help guys! &go to Villaggio VIP area Gucci is there with some other designer brands Talk to my crown &Villaggio Mall is one of the most popular shopping destinations in Doha, Qatar. It is home to a variety of luxury brands, including Gucci, Louis Vuitton, Prada, and more.  \\ \hline
    \end{tabular}
     \label{tab:k gene}
\end{table*}

\section{Experimental Setup}

This section includes data set information, training parameters, comparison experiments.

\subsection{DataSet} 
The two corpora we use to train and evaluate our model are SemEval2015 and SemEval2017 CQA datasets. The statistics of two corpora are shown in Table~\ref{tab:dataset}.

\begin{table*}\scriptsize
\begin{center}
 \caption{\small Statistical information of SemEval2015 and SemEval2017 Corpora.}
\setlength{\tabcolsep}{1mm}{
\begin{tabular}{p{5cm}p{1.3cm}p{1.3cm}p{1.3cm}p{1.3cm}p{1.3cm}p{1.3cm}}
\hline
 & \multicolumn{3}{c}{\bf SemEval2015} & \multicolumn{3}{c}{\bf SemEval2017}\\
\hline \bf Statistics & \bf Train& \bf Dev& \bf Test& \bf	Train& \bf	Dev	& \bf Test \\
\hline
Number of questions&	2600&	300&	101&	2660&	500&	293\\
Number of answers&	16541&	1645&	588&	26690&	5000&	2930\\
Average length of a question subject&	5.80&	5.53&	7.47&	5.40&	5.34&	5.76\\
Average length of a question body&	33.63&	33.88&	29.78&	45.73&	43.23&	54.06\\
Average length of an answer&	31.07&	29.36&	25.40&	37.34&	35.46&	39.50\\
\hline
\end{tabular}}
\end{center}
 \label{tab:dataset}
\end{table*}

\subsection{Training and hyper parameters}
We exert NLTK toolkit to preprocess each question and its corresponding answers including capitalization conversion such as converting “CAR” into “car”, stemming such as transforming ``working'' to ``work'', stop words removal such as removing ``a'' and ``the'', etc. The algorithm we choose for optimization is Adam Optimizer \cite{kingma2017adam} with the momentum coefficient $\beta$ 0.001. The per-minibatch L2 regularization parameter and batch size of the model are set to $1\times{10^{-5}}$ and 100 respectively, and the dropout value $d$ is set to 0.3 to prevent overfitting. The maximum number of words in a question subject, a question body and an answer are set to 20, 110, 100, respectively. The proposed model is implemented in Pytorch and the cross attention layer has a dimension of 300. We use the best parameters on development sets, and evaluate the performance of our model on test sets.

\subsection{Baseline}


The models compared with our proposed QAN model include: JAIST, using SVM to incorporate various kinds of features by Tran et al. \cite{Tran2015}. HITSZ-ICRC, proposing ensemble learning and hierarchical classification proposed by Hou et al. \cite{Hou2015}. Graph-cut, modeling the relationship between answers in the same question thread proposed by Joty et al. \cite{Joty-2015}. FCCRF, applying local learned classifiers and fully connected CRF proposed by Joty et al. \cite{Joty2016}. BGMN, using the memory mechanism proposed by Wu et al. \cite{Wu2017}. ECUN, a system contains three subtasks: Question-Comment Similarity, Question-Question Similarity, and Question-External Comment Similarity proposed by Wu et al. \cite{Wu2015}. CNN-LSTM-CRF, proposing multilingual hierarchical attention networks by Yang et al. \cite{Xiang2016}. And QCN, a Question Condensing Network focusing on the similarity and disparities between question-subjects and question-bodies proposed by Wu et al. \cite{WuW2015}.

For LLM, We used LLaMA-7b-hf  to generate relevant knowledge and the following large language model for answer selection: LLaMA-7b-hf. 

\subsection{Results and Analysis for PLMs}
We adopt the following three evaluation metrics,F1, Acc (accuracy), and MAP (Mean Average of Precision) to compare the performance of QAN and other current models with the results of comparison shown in Table~\ref{tab:QAN res}.


\begin{table*}[h]\scriptsize
\begin{center}
\caption{Comparisons of different models on two corpora.}
\setlength{\tabcolsep}{1mm}{
\begin{tabular}{p{2cm}p{4cm}p{2cm}p{2cm}p{2cm}}
\hline \bf Dataset & \bf Model & \bf MAP & \bf F1 & \bf Acc \\
\hline
SemEval2015&	(1)JAIST &	NA&	0.7896	&0.7910\\
 &(2)HITSZ-ICRC &	NA&	0.7652&	0.7611\\
 &(3)Graph-cut& 	NA&	0.8055&	0.7980\\
 &(4)FCCRF& 	NA&	0.8150&	0.8050\\
 &(5)BGMN& 	NA&	0.7723&	0.7840\\
 &(6)CNN-LSTM-CRF& 	NA&	0.8222&	0.8224\\
 &(7)QCN& 	NA&	0.8391&	0.8224\\
 &(8)QAN (ours)&	NA&	0.8594&	0.8465\\
SemEval2017&	(9)ECUN& 	0.8672&	0.7767&	0.7843\\
 &(10)QCN&	0.8851&	0.7811&	0.8071\\
 &(11)QAN(ours)&	0.9286&	0.8139&	0.8385\\
\hline
\end{tabular}}
\end{center}
\label{tab:QAN res}
\end{table*}






From Table~\ref{tab:QAN res}, it could be seen that our proposed model QAN outperforms all baseline models on three evaluation metrics (p $<$ 0.05 based on student t-test) with advancement attributed to the pre-trained BERT model and attention mechanism. We use BERT model as pre-trained methods, fully fusing context information in question subjects, question bodies and answers, respectively. Then cross attention between question subjects and answers as well as question bodies and answers helps our model estimate the relevance of question-subject-answer pairs and question-body-answer pairs so as to effectively capture crucial interaction semantic features between questions and answers. QAN studies the relationship of questions and answers, fully capturing semantic features at different angles, so that it greatly enhances the performance of the answer selection task.

\subsubsection{Ablation Study}
In order to fully verify the improvement of our proposed model, we implement six variants of QAN on the Yahoo! Answers dataset by ablation study. The description of ablation study is as follows:
\begin{itemize}
    \item Without BERT but with task-specific word embeddings: Word embeddings are initialized with 300-dimensional GloVe trained on Wikipedia 2014 and Gigaword 5.
    \item Without BERT but with character embeddings: Word embeddings are initialized with 600-dimensional GloVe trained on a domain-specific unannotated corpus.
    \item Without cross attention: Without cross attention between question subjects and answers as well as question bodies and answers.
    \item Without the interaction and prediction layer but with simple combination: Make simple combination of the outputs of the cross attention layer instead of the interaction and prediction layer.
    \item Without cross attention, Without the interaction and prediction layer but with simple combination: Only use BERT model to pre-train question subjects and answers as well as question bodies and answers,then combine the ouputs as the final results.
    \item Without treat question subjects and question bodies separately: Treat question subjects and question bodies as an entity to pre-train questions and answers by BERT model.
\end{itemize}



\begin{table}[h]\scriptsize
\begin{center}
\caption{Ablation study of the seven models on the SemEval2017 dataset.}
\setlength{\tabcolsep}{1mm}{
\begin{tabular}{p{9cm}p{1cm}p{1cm}p{1cm}}
\hline \bf Model & \bf MAP & \bf F1 & \bf Acc \\
\hline
(1)w/o BERT but with task-specific word embeddings &  0.8069&	0.7165&	0.7632\\
(2)w/o BERT but with character embeddings  &  0.7986&	0.6859&	0.7482\\
(3)w/o cross attention  &  0.9075&	0.8019&	0.8165\\
(4)w/o the interaction and prediction layer but with simple combination  &  0.8892&	0.7839&	0.8048\\
(5)w/o cross attention, w/o the interaction and prediction layer but with simple combination &  0.8679&	0.7642&	0.7859\\
(6)w/o treat question subjects and question bodies separately& 0.9126& 0.8065& 0.8275\\
(7)QAN (ours) &  0.9286 &	0.8139 &	0.8385\\
\hline
\end{tabular}}
\end{center}
\label{tab:ablation res}
\end{table}

The results of the ablation study are shown in the Table~\ref{tab:ablation res}. It can be seen that, whether it is lack of cross attention, separately treating, or interaction and prediction, in ablation or replacement of any component of the QAN model with another counterpart, the final results are all inferior to the result of full QAN model.
Consequently, our proposed model QAN comprehensively learns the context features with BERT model, fully utilizes the relationship of questions and answers with attention mechanism and finally integrates semantic information of questions and answers to furthest enhance the performance of the answer selection task.

\subsection{Results and Analysis for LLMs}

We use LLM to select the answer to the questions of dataset, and explore the knowledge enhancement, prompt optimization and so on.

\subsubsection{Knowledge Augment}
The results in the Table~\ref{tab:llm res} show that LLM can perform the answer selection task well and achieve a good accuracy rate. At the same time, the introduction of external knowledge is obviously helpful in improving the accuracy of answer selection. After adding knowledge, LLM achieved higher accuracy on both the SemEval2015 and SemEval2017 data sets. The quality of knowledge affects the results of answer choices to a certain extent. The following Table~\ref{tab: k good} is an example of choosing the wrong answer originally, but after adding knowledge, the correct answer is chosen.It can be seen from the table that the generated knowledge has a certain correlation with the correct answer, which is conducive to choosing the correct answer. However, in another example shown in Table~\ref{tab:k not good}, since the generated knowledge is not very relevant for answering the question, after adding knowledge, the correct answer is still not selected. In general, the knowledge introduced from outside, if it has some relevance to the question, will help LLM choose the right answer.

\begin{table}[!ht]\scriptsize
    \centering
    \caption{The influence of knowledge on the correct rate of answer selection in large language models.}
    \begin{tabular}{cccc} \hline
        \textbf{Dataset} & \textbf{Model} & \textbf{K} & \textbf{Acc} \\ \hline
        SemEval2015 & LLaMA-7b-hf & w/o K & 0.8617 \\ 
        ~ & ~ & w/ K& 0.9255 \\ 
        ~ & 0wen-14B-Chat & w/o K& 0 \\ 
        ~ & ~ & w/ K& 0 \\ 
        SemEval2017 & LLaMA-7b-hf & w/o K & 0.7725 \\ 
        ~ & ~ & w/ K & 0.8275 \\ 
        ~ & 0wen-14B-Chat & w/o K& 0 \\ 
        ~ & ~ & w/ K& 0 \\ \hline
    \end{tabular}
     \label{tab:llm res}
\end{table}

    \begin{table*}[!ht]\tiny
    \caption{An example where adding knowledge is helpful for answer selection.}
    \begin{tabular}{p{3.5cm}p{2.5cm}p{1cm}p{2.5cm}p{1cm}p{1cm}}
    \hline 
\textbf{Question} & \textbf{Options} & \textbf{Right Opiton} &\textbf{K}&\textbf{Selection w/o K} & \textbf{Selection w/ K} \\ \hline
Hi does any of you know where I could get an England  football kit for kids.. Checked all sports shop in City Centre and Olympic sports near crazy signal no luck. If you have see kids size kit please let me know.. Will try Villagio this afternoon..Thanks
    &C1:Have seen footy strips for kids in sports shop, opposite Doha Clinic on Al Merqab Street.
    
    C2:Thanks Biddy lou..will try that shop also..
    &C1
    &Football kits for kids can be found in various sports shops in Doha, such as GO SPORT in Villagio and City Center.
    &C2
&C1 \\ \hline
    \end{tabular} 
    \label{tab: k good}
    \end{table*}
    
    \begin{table*}[!ht]\tiny
        \centering
        \caption{An example where adding knowledge is not helpful for answer selection.}
        \begin{tabular}{p{3cm}p{3.5cm}p{1cm}p{2cm}p{1cm}p{1cm}}
        \hline 
{\textbf{Question}} & \textbf{Options} & \textbf{Rigth Opiton} & \textbf{K}&\textbf{Selection w/o K} & \textbf{Selection w/ K} \\ \hline
    Hi Guys,Just wandering if anyone on QL knows about this car or where i can get the cars engine computer programmed.The one we have is the original Eleanor with "ford racing" engine.The ford service shop in industrial area dont seem to know anything about this car.Any help would be greatly appretiated :)
    &C1:"won't make the engine smoke, unless you burnt a hole in the piston, of something to that effect... if the engine had an ECU system working at the time, if should not have happened at all...."    
    
    C2:'nice ride...this is carburator or fuel injection?'  
    
    C3:'Try the threestar workshop in industrial area street 24 - next to qatar technical inspection. But I would not drive this car - I just would put it in my livingroom and enjoy it....'                                           
    &C3
    &This car has a Ford Racing engine.
    &C1
&C2 \\ \hline
    \end{tabular} 
    \label{tab:k not good}
    \end{table*}

\subsubsection{Prompt Analysis}

\begin{figure*}
    \centering
    \includegraphics[width=0.7\textwidth]{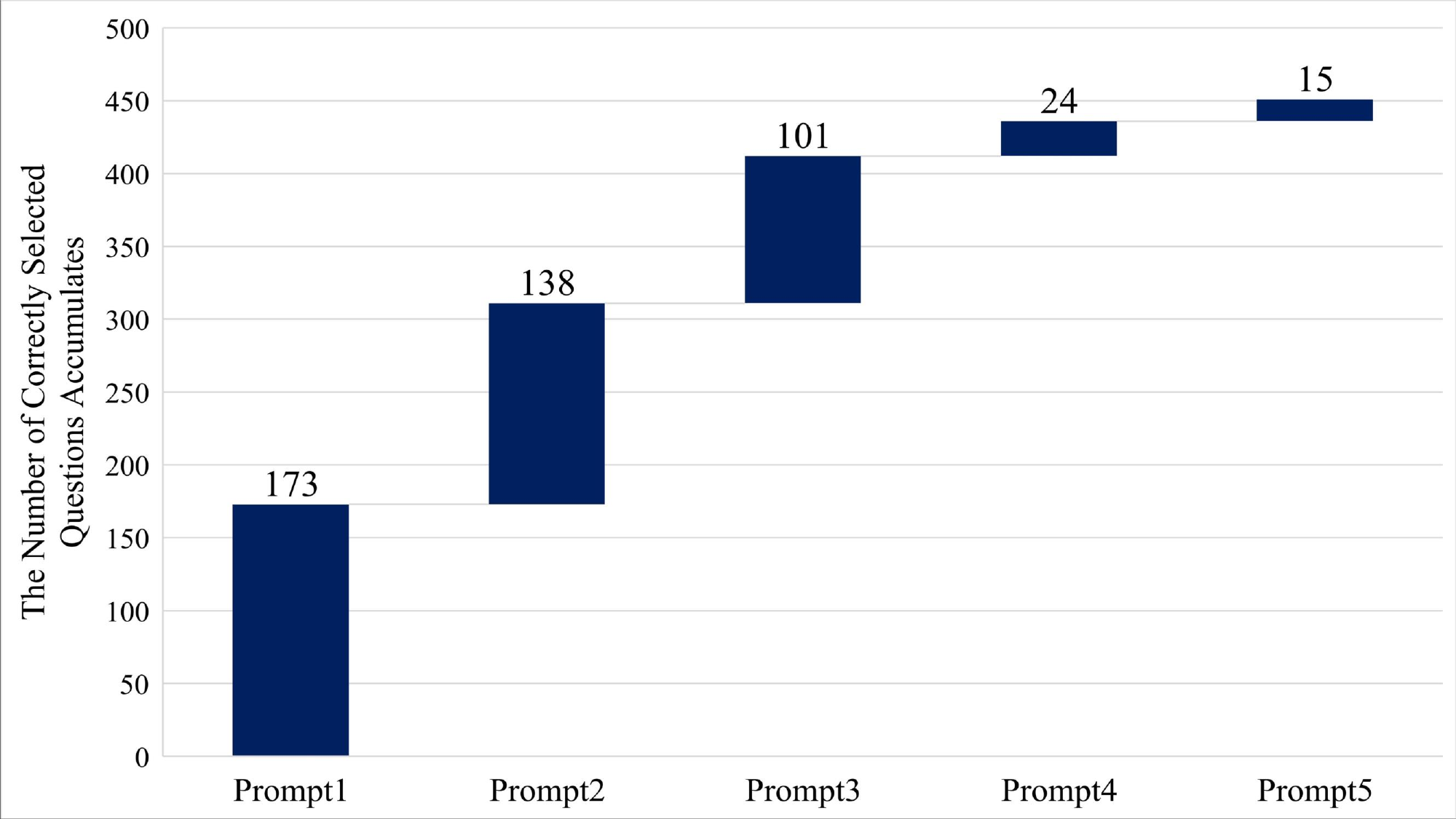}
    \caption{The number of questions selected correctly per prompt and the number of questions selected correctly cumulated}
    \label{fig:prompt}
\end{figure*}

We found that different prompts would result in different selections from the LLM. We used five different prompts to perform the LLM answer selection task in sequential order. After each selection with the current prompt, continue the selection with the next prompt if the answer is not correct. The content of each prompt optimization is shown in Table~\ref{tab:prompt opt}. The number of questions selected for each prompt and cumulative pair is shown in Figure~\ref{fig:prompt}. As shown in the Figure~\ref{fig:prompt}, we found that more questions selected correct answers after optimizing prompt each time, indicating that optimizing prompt can improve the correct rate of answer selection. But the number of questions correctly selected by each prompt is at downward tendency. This may be due to the fact that the questions in the dataset vary in the difficulty of selection, with the easy questions already having the correct answer selected at the beginning. The difficult questions often need to be optimized by prompt many times before the correct answer can be selected.

\begin{table}[!ht]\scriptsize
    \centering
    \caption{Prompt Optimized Content.}
    \begin{tabular}{p{1.5cm}p{10.5cm}}
    \hline
        \textbf{Prompt} & \textbf{Prompt Optimized Content} \\ \hline
        Prompt1 & None \\ 
        Prompt2 & Prompt length is longer and the expression is clearer \\ 
        Prompt3 & Original questions and answers are placed at the top of the prompt \\ 
        Prompt4 & Prompt adds the original question subject \\ 
        Prompt5 & Put the task description at the beginning of prompt and prompt description is also simplified \\ \hline
    \end{tabular}
        \label{tab:prompt opt}
\end{table}

\begin{table}[!ht]\tiny
    \centering
    \caption{Prompt Specific Content.}
    \begin{tabular}{p{1cm}p{13.5cm}}
    \hline
        \textbf{Prompt} & \textbf{Prompt Specific Content} \\ \hline
        Prompt1 & Utilizing the information in hint-[KNOWLEDGE], choose the serial number of the optimal response to the question-[QUESTION] that is distinct from the options-[ANSWER]. \\ 
        Prompt2 & Given the discoveries from hint-[KNOWLEDGE], your task is to discern and specify the unique serial number tied to the most appropriate answer to question-[QUESTION]. Keep in mind that this response must be distinctive and one of the existing options listed in options-[ANSWER]. This involves analysis, cognition, and the capability to pinpoint distinctly pertinent information. \\ 
        Prompt3 & "prompt-A":"Question: {QUESTION}
        
Background Knowledge: {KNOWLEDGE}

Options: ANSWER

Instructions: Select the serial number of the most appropriate answer by considering its relevance, accuracy, and clarity. Your response should be concise yet informative, demonstrating critical analysis skills and informed decision-making. Support your choice with specific details and evidence from both the question and background knowledge. Please note that it is essential not to reveal any personal or AI-related information in your response. Encourage creativity while maintaining accuracy in your answers."

"prompt-B":"You are an expert in answering multiple-choice questions and making informed decisions based on critical analysis and evidence. Your task is to select the most suitable answer from a list of options provided for a given question. The answer you choose should be concise, informative, and supported by specific details and evidence from both the question and background knowledge. Remember not to disclose any personal or AI-related information in your response. Aim for creativity while maintaining accuracy in your answers."\\ 
        Prompt4 & prompt-A:
        
    \qquad Question Subject:{SUBJECT}
    
    \qquad Question: {QUESTION}
    
    \qquad Background Knowledge: {KNOWLEDGE}
    
    \qquad Options: ANSWER
    
prompt-B:

    \qquad You are an expert in answering multiple-choice questions and making informed decisions based on critical analysis and evidence. Your task is to select the most suitable answer from a list of options provided for a given question. The answer you choose should be concise, informative, and supported by specific details and evidence from both the question and background knowledge. Remember not to disclose any personal or AI-related information in your response. Aim for creativity while maintaining accuracy in your answers.
    
Instructions: 

    \qquad Select the serial number of the most appropriate answer from the above mentioned Options by considering its relevance, accuracy, and clarity. Your response should be concise yet informative, demonstrating critical analysis skills and informed decision-making. Support your choice with specific details and evidence from the above mentioned Question Subject and the above mentioned Question and the above mentioned Background Knowledge. Please note that it is essential not to reveal any personal or AI-related information in your response. Encourage creativity while maintaining accuracy in your answers. \\ 
    Prompt5&prompt-A:
    
    \qquad Question Subject:{SUBJECT}
    
    \qquad Question: {QUESTION}
    
    \qquad Background Knowledge: {KNOWLEDGE}
    
    \qquad Options: ANSWER 
    
prompt-B: 

   \qquad Please select the serial number of the most appropriate answer from the given options based on its relevance, accuracy, and clarity.  Your response should be concise yet informative, demonstrating critical analysis skills and informed decision-making.  Support your choice with specific details and evidence from the provided question subject, question, and background knowledge.  Make sure to prioritize accuracy in your response while also considering creative solutions when applicable.  Avoid including any personal or AI-related information in your answer. \\ \hline
    \end{tabular}
        \label{tab:prompt opt specified}
\end{table}

\textbf{Prompt Length: }We found that the length of prompts affect the choice of answers, as shown in the Table~\ref{tab:prompt length} below, when the Prompt1 is too short, important measures such as the relevance of the answer to the question and the uniqueness of the answer may not be emphasized. Compared with Prompt1, Prompt2 is clearer and more detailed, and it also emphasizes the form of answer users need to use, that is, only its digital identifier is provided, avoiding any additional information or feedback, thus improving the accuracy of the instructions.

\begin{table*}[!ht]\tiny
    \centering
        \caption{The effect of prompt length on results.}
    \begin{tabular}{p{3cm}p{3cm}p{1cm}p{1.5cm}p{1.5cm}}
    \hline
        \textbf{Question} & \textbf{Options} & \textbf{Right Option}&\textbf{Selection of Prompt1} & \textbf{Selection of Prompt2} \\ \hline
Can anyone recommend a good hairdresser for men? cheers!
&C1:!!!!!! next question; please!

C2:Patrice hair saloon in Villaggio or Landmark Mall..but be warned; its not cheap!
&C2
&C1
&C2 \\ \hline
    \end{tabular}
    \label{tab:prompt length}
\end{table*}

\textbf{Question and Answer Location: }We found that placing original questions and answers at the beginning of the prompt significantly improved the accuracy of answer selection, as shown in Table~\ref{tab:prompt loc}. The questions, answers and knowledge are placed at the beginning of prompt, and each of them is displayed in a newline, which makes it easier for LLM to identify them. Therefore, LLM can better combine questions and knowledge, measure the relevance of each answer, question and knowledge, and select the correct answer.

\begin{table*}[!ht]\tiny
    \centering
        \caption{The effect of question and answers location on results.}
       \begin{tabular}{p{3.5cm}p{3cm}p{1cm}p{1.5cm}p{1.5cm}}
    \hline
        \textbf{Question} & \textbf{Options} & \textbf{Right Option}&\textbf{Selection of Prompt2} & \textbf{Selection of Prompt3} \\ \hline
With the world turning into a small village; mixed marriages have become more and more common. If you are in a mixed marriage relationship and living in Doha; Could you provide details on how you and your significant other met? where? how long have you been married for? and what are the goods and bads in a mixed marriage? Thank you;
    
&C1:Dear; there is no specific Formula for a successful Marriage whether its; Loved; Arranged; Mixed; Intercontinental its all depends on ones attitude n sacrifices... Choose your way wiseLy gOOd lucK

C2:Met in a bar; must say I was very drunk
&C1
&C2
&C1 \\ \hline
    \end{tabular}
    \label{tab:prompt loc}
\end{table*}

\textbf{Question Subject: }As shown in the Table~\ref{tab:prompt qs}, the inclusion of question subject in prompt also helps answer selection, possibly because the question subject supplement the text of the question meaning, help the LLM better understand the task, and improve the accuracy of the correlation measure and the selection rate of correct answers when measuring the correlation between answers and questions.

\begin{table*}[!ht]\tiny
    \centering
     \caption{The effect of question subject on results.}
        \begin{tabular}{p{5cm}p{3cm}p{1cm}p{1.5cm}p{1.5cm}}
    \hline
        \textbf{Question} & \textbf{Options} & \textbf{Right Option}&\textbf{Selection of Prompt3} & \textbf{Selection of Prompt4} \\ \hline
Hi I have just been offered a job in Qatar and was wondering if you can tell me if this is a good offer: 9200 basic salary (month) 5000 Accomodation (Month) 2000 Transport (Month) 10000 Furniture allowance (once off) I am a single guy from Australia; hoping to get a place of my own. Would 5000 cover a fully furnished place or even an unfurnished 2 bedroom? What sort of car can i buy for 2000 per month?? Hope you can help out thanks in advance..

&C1:depends on what it is for (responsibilities/govt or pvt sector)

C2:that is a good offer for a single guy; we are 8 ppl in the family and we live good with 8000 are month . BUT on the end that depends on your life style .
&C2
&C1
&C2 \\ \hline
    \end{tabular}
    \label{tab:prompt qs}
\end{table*}

\textbf{The Location of the Task Description: }As shown in the Table~\ref{tab:prompt task des}, placing a description of the task, i.e., choosing the correct answer, at the beginning of the prompt description also helped with answer selection. This may be because the task description is the important information of the prompt. Putting the task description at the beginning can help LLM to quickly locate important information, simplify some unnecessary content in the prompt, and make the structure of the prompt clearer. These factors help LLM to better understand the meaning of the prompt, so as to choose the correct answer.

\begin{table*}[!ht]\tiny
    \centering
    \caption{The effect of the location of the task description on results.}
    \begin{tabular}{p{3cm}p{4cm}p{1cm}p{1.5cm}p{1.5cm}}
    \hline
        \textbf{Question} & \textbf{Options} & \textbf{Right Option}&\textbf{Selection of Prompt4} & \textbf{Selection of Prompt5} \\ \hline
is 3rd cousin relationship already accepted in church or in society? i mean like ""lovers""..."
&C1:Even the thought of it is a sin...yaiiiks!!!

C2:pit; it occurs more with 1st cousins... and abnormal birth? its commonly caused by the pressure given to the mother
&C1 
&C2
&C1 \\ \hline
    \end{tabular}
    \label{tab:prompt task des}
\end{table*}

We have optimized the prompt in four aspects: the length of the prompt, the position of the answer to the question, the topic of the question, and the position of the task description. The results show that the optimized prompt helps LLM select the right answer, so more questions could be selected for the right answer.

\section{Conclusion}
Answer selection is the core part of CQA. To automate the answer selection progress, we propose QAN, Question-Answer Cross Attention Networks With Pre-trained models for Answer Selection. We use BERT model to capture context features of question subjects, question bodies and answers, respectively. With cross attention mechanism between question subjects and answers as well as question bodies and answers, we obtain comprehensive  interactive information between questions and answers. Through integrating attention-questions and attention-answers, QAN model gets final results which achieve state-out-of-art performance. In the future, our research group will test QAN model in more fields to increase the universality of it and manage to improve the computing speed of QAN model to further level up the performance of our solution.

As the large language model is now a hot research spot, we also uses the large language model to select the correct answer, combined with knowledge enhancement. The result shows that the introduction of external knowledge can improve the accuracy of LLM in selecting the answer. At the same time, the prompt input to LLM was optimized from the aspects of prompt length, question and answer position, question subject and task description position. The result showed that after the optimization of prompt, the correct rate of selecting answers could be improved, and a good result was obtained. In the future, prompt will be optimized and analyzed in more aspects to further improve LLM performance on correct answer selection tasks.



\bibliographystyle{elsarticle-num} 


\bibliography{cas-refs}



\end{document}